\documentclass[conference]{IEEEtran}
\IEEEoverridecommandlockouts
\usepackage{cite}
\usepackage{amsmath,amssymb,amsfonts}
\usepackage{algorithmic}
\usepackage{graphicx}
\usepackage{textcomp}
\usepackage{booktabs}
\usepackage{balance}
\usepackage{pifont}
\usepackage[dvipsnames]{xcolor}
\usepackage{url}

\usepackage{tikz}
\usetikzlibrary{backgrounds}
\usetikzlibrary{arrows,shapes}
\usetikzlibrary{tikzmark}
\usetikzlibrary{calc}
\usepackage{blindtext}
\usepackage{tcolorbox}
\usepackage{tikz}
\usetikzlibrary{arrows,shapes,positioning,shadows,trees,mindmap}
\usepackage[edges]{forest}
\usetikzlibrary{arrows.meta}
\colorlet{linecol}{black!75}
\usepackage{xkcdcolors} 
\usepackage{tikz}
\usetikzlibrary{backgrounds}
\usetikzlibrary{arrows,shapes}
\usetikzlibrary{tikzmark}
\usetikzlibrary{calc}
\newcommand{\highlight}[2]{\colorbox{#1!17}{$\displaystyle #2$}}

\renewcommand{\highlight}[2]{\colorbox{#1!17}{#2}}

\def\W{{\mathbf W}}
\def\x{{\mathbf x}}
\def\ii{{\hat{\imath}}}												
\def\ij{{\hat{\jmath}}}												
\def\ik{{\hat{\kappa}}}												

\def\L{{\mathcal L}}
\def\Exp{{\mathbb E}}

\begin{document}

\title{Hypercomplex Image-to-Image Translation}

\author{\IEEEauthorblockN{Eleonora Grassucci, Luigi Sigillo, Aurelio Uncini, and Danilo Comminiello\thanks{This research was funded by "Progetti di Ricerca" of Sapienza University of Rome under grant numbers RM120172AC5A564C and RG11916B88E1942F.}}
\IEEEauthorblockN{\textit{Dept. Information Engineering, Electronics and Telecommunications (DIET), Sapienza University of Rome, Italy}\\
Email: eleonora.grassucci@uniroma1.it.}}

\maketitle

\begin{abstract}
Image-to-image translation (I2I) aims at transferring the content representation from an input domain to an output one, bouncing along different target domains. Recent I2I generative models, which gain outstanding results in this task, comprise a set of diverse deep networks each with tens of million parameters. Moreover, images are usually three-dimensional being composed of RGB channels and common neural models do not take dimensions correlation into account, losing beneficial information. In this paper, we propose to leverage hypercomplex algebra properties to define lightweight I2I generative models capable of preserving pre-existing relations among images dimensions, thus exploiting additional input information. On manifold I2I benchmarks, we show how the proposed Quaternion StarGANv2 and parameterized hypercomplex StarGANv2 (PHStarGANv2) reduce parameters and storage memory amount while ensuring high domain translation performance and good image quality as measured by FID and LPIPS score. Full code is available at \url{https://github.com/ispamm/HI2I}.
\end{abstract}


\section{Introduction}
\label{sec:intro}

The aim of image-to-image translation (I2I) is to learn a function $G_{S \rightarrow T}$ able to generate an image $x_{ST}$ by translating an input image $x_S \in S$ from the source domain $S$ to a target domain $T$. More formally,

\begin{equation}
\label{eq:i2i}
    x_{ST} = G_{S \rightarrow T}(x_S), \qquad x_{ST} \in T,
\end{equation}

\noindent in which the target domain can be injected by learning domain features from a reference image or by sampling latent vectors from the domain space \cite{Choi2020StarGAN2}.

Recently, I2I applications are becoming widespread, including a plethora of diverse tasks such as attribute manipulation \cite{He2019AttGAN, Choi2018StarGAN, Choi2020StarGAN2, Lin2021TPAMI}, sketch-to-image \cite{Almahairi2018CycleGAN, Zhang2020CrossDomainCL}, style transfer \cite{Kim2020UGATIT, Lee2020drit}, semantic synthesis \cite{Park2019SPADE, Chen2020Dist}, and others \cite{Wu2021StyleSpace, chong2021gans, karras2021aliasfree, alaluf2022times}. Among these, generative adversarial networks (GANs) \cite{GoodfellowNIPS2014} are particularly suitable for this task due to few restrictions on the generator network. Indeed, GANs are deep generative models composed of a network whose goal is to generate new images given an input noise vector and a discriminator network that aims at distinguishing generated images from real one. These are trained in an adversarial fashion requiring very few constraints on models architecture, especially on the generator. The latter network covers the role of the mapping function $G_{S \rightarrow T}$ in Eq.~\ref{eq:i2i} for GAN-based I2I applications.

Latest works on generative models, including I2I ones, have achieved impressive results by scaling-up models in terms of trainable parameters, computational complexity and memory requirements \cite{karras2020analyzing, Brock2019LargeSG, schonfeld2021unet}. Therefore, most of them are difficult to train with a lower budget, undermining their replicability. Furthermore, low attention has been paid to how multidimensional inputs such as color images are processed by these models. The human eye perceives an image with lots of color shades that are the result of interactions among the three RGB channels. Therefore, channels interplays are crucial for a proper image processing. Actually, common real-valued models do not leverage this detail treating each channel as a separate entity, causing an information loss.

To overcome these limitations, quaternion neural networks (QNNs) have been proposed. Such models have recently gained increasing attention due to their ability to exploit quaternion algebra properties that lead to a consistent parameters reduction while ensuring high performance \cite{ParcolletAIR2019, VALLE2021111, ComminielloICASSP2019a, Brignone2022ISCAS}. This is because, despite the lower number of parameters due to the vector multiplication which, in this domain, takes a different form with respect to the real-valued domain, QNNs preserve correlations among channels. Indeed, they process RGB images as a single entity, thus exploiting relations within input dimensions, building a more suitable representation \cite{ParcolletICASSP2019a}. Both GANs and variational autoencoders (VAEs) have been defined in the quaternion domain demonstrating encouraging scores and an improved generation ability \cite{GrassucciQGAN2021, GrassucciICASSP2021, Grassucci2021Entropy}. However, QNNs accept $4$D inputs only, thus a padding channel has to be annexed to RGB images to compose a pure quaternion. This useless information may undermine QNNs performance.

Lately, novel approaches have been proposed to extend QNNs advantages to any $n$D input \cite{Zhang2021PHM, grassucci2021lightweight, mahabadi2021compacter, le2021parameterized}. Thanks to a parameterized sum of Kronecker products, these methods learn hypercomplex algebra regulations directly from data. Thus, they can be defined in any $n$D domain of user choice without needing pre-fixed multiplication rules as it is instead for QNNs. 

In this paper, we propose to exploit hypercomplex algebra properties to define lightweight and replicable models for multi-domain I2I tasks. Therefore, we introduce a Quaternion StarGANv2 and the parameterized hypercomplex StarGANv2 (PHStarGANv2), a novel element in the family of parameterized hypercomplex neural networks (PHNNs). The PHStarGANv2 is able to behave like the Quaternion StarGANv2 by involving $n=4$ or to process RGB images in their natural domain with $n=3$. These models save trainable parameters and storage memory, thus are more accessible with a lower budget. Moreover, thanks to hypercomplex vector multiplications, including the Hamilton product, the Quaternion StarGANv2 and the PHStarGANv2 grasp relations among image channels thus learning more information despite the lower number of parameters. We prove these abilities with an empirical evaluation on multiple benchmarks and evaluate the performance with visual inspections and different objective metrics.
Accordingly, our contributions follow.
\begin{enumerate}
    \item We introduce a set of hypercomplex StarGANv2 for image-to-image translation, including the Quaternion StarGANv2 and the parameterized hypercomplex StarGANv2 (PHStarGANv2), which leverage hypercomplex algebra properties to reduce the overall number of parameters and exploiting channels correlations. The latter leads to better performance in terms of generation quality and translation accuracy. To the best of our knowledge, this is the first generative model defined with parameterized hypercomplex layers.
    \item We define the instance normalization for hypercomplex domains. This novel technique increases performance of quaternion and parameterized hypercomplex models as measured by FID score.
    \item We propose a novel method to initialize parameterized hypercomplex layers that avoids training degeneracy, which is due to the almost-zero values of the weight matrix. We show how weights are distributed more properly with our approach and that this leads to a crucial improvement in generation results, as measured by objective metrics.
\end{enumerate}

The rest of the paper is organized as follows. Section~\ref{sec:qnn} introduces theoretical concepts of quaternion generative models and parameterized hypercomplex networks. In Section~\ref{sec:method}, we expound the proposed I2I models, in Section~\ref{sec:qinst} we conceive the hypercomplex normalizations, while in Section~\ref{sec:init} we introduce the novel inizialization. Then, Section~\ref{sec:exp} proves the experimental validity of our methods. Finally, conclusions are drawn in Section~\ref{sec:conc}.

\section{Quaternion and Hypercomplex Generative Models}
\label{sec:qnn}

Quaternion and hypercomplex neural networks were born from a system number based on a set of hypercomplex numbers $\mathbb{H}$, whose additions and multiplications are regulated by a collection of algebra rules. Among these numbers, quaternions are identified by three imaginary units, namely $\ii, \ij$ and $\ik$, and four real-valued coefficients $q_c, \; c=\{0,1,2,3\}$ as:
\begin{equation}
    q = q_0 + q_1 \ii + q_2 \ij + q_3 \ik.
\end{equation}

While the addition of two quaternions $p$ and $q$ is intuitive, being $p+q = (p_0+q_0) + (p_1+q_1)\ii + (p_2+q_2)\ij + (p_3+q_3)\ik$, a more detailed formula is needed to model imaginary units interplays when vector multiplication is performed. Indeed, this operation is not commutative in the quaternion domain since $\ii\ij = - \ij\ii, \; \ii\ik = - \ik\ii, \; \ij\ik = - \ik\ij$, and so on. To this end, for a proper quaternion vector multiplication, the Hamilton product has been introduced. The Hamilton product is the core of quaternion neural networks (QNNs), where a quaternion weight matrix $\W = \W_0 + \W_1\ii + \W_2\ij + \W_3\ik$ is multiplied by a quaternion input $\x$ with the same structure as:

\begin{equation}
{\bf{W}} \otimes {\bf{x}} = \left[ {\begin{array}{*{20}c}
   \hfill {{\bf{W}}_0 } & \hfill { - {\bf{W}}_1 } & \hfill { - {\bf{W}}_2 } & \hfill { - {\bf{W}}_3 } \\
   \hfill {{\bf{W}}_1 } & \hfill {{\bf{W}}_0 } & \hfill { - {\bf{W}}_3 } & \hfill {{\bf{W}}_2 } \\
   \hfill {{\bf{W}}_2 } & \hfill {{\bf{W}}_3 } & \hfill {{\bf{W}}_0 } & \hfill { - {\bf{W}}_1 } \\
   \hfill {{\bf{W}}_3 } & \hfill { - {\bf{W}}_2 } & \hfill {{\bf{W}}_1 } & \hfill {{\bf{W}}_0 } \\
\end{array}} \right] \otimes \left[ {\begin{array}{*{20}c}
   {{\bf{x}}_0 } \hfill  \\
   {{\bf{x}}_1 } \hfill  \\
   {{\bf{x}}_2 } \hfill  \\
   {{\bf{x}}_3 } \hfill  \\
\end{array}} \right].
\label{eq:qconv}
\end{equation}

\noindent Weight submatrices are reused and shared among input components, thus while the dimension of the matrix $\W$ is the same as a real-valued one, it involves just $1/4$ parameters of its real-valued counterpart. Furthermore, by sharing weights among different input dimensions, quaternion layers exploit correlations contained within components thus preserving the original multidimensional structure of the input while gaining advantages from it. This ensures good performance despite the lower number of trainable parameters.
Nevertheless, QNNs are limited to $4$D data, thus when processing RGB images with three channels, an uninformative further channel has to be padded in order to build the four-dimensional input.

Recently, a novel approach for parameterizing hypercomplex models has been proposed \cite{Zhang2021PHM, grassucci2021lightweight}. It aims at building the hypercomplex weight matrix as a sum of Kronecker products parameterized by a user-defined hyperparameter $n$. These methods allow the definition of fully parameterized hypercomplex neural networks (PHNNs) that completely work in the chosen hypercomplex domain. More in detail, the weight matrix $\mathbf{H}$ of a generic PHNN is defined as

\begin{equation}
\label{eq:phc}
    \mathbf{H} = \sum_{i=1}^n \mathbf{A}_i \otimes \mathbf{F}_i,
\end{equation}

\noindent whereby $\mathbf{A}_i$ describe the hypercomplex algebra rules by learning them directly from data (i.e., the Hamilton product for the quaternion domain) and $\mathbf{F}_i$ are batch of weights that can be scalars for fully connected (FC) layers, or groups of filters for convolutional ones. In the first case, we deal with parameterized hypercomplex multiplication (PHM) layers \cite{Zhang2021PHM, mahabadi2021compacter}, while in the second one we employ parameterized hypercomplex convolutional (PHC) layers \cite{grassucci2021lightweight}.

PHNNs involve $1/n$ free parameters of their real-valued counterparts while being more efficient than quaternion models and obtaining better results due to the fully-learnable structure of their layers. By fixing $n=4$, through Eq.~\ref{eq:phc}, we can express the Hamilton product in Eq.~\ref{eq:qconv}. PHNNs exceed QNNs thanks to their ability to grasp a more suitable weights organization (i.e., the rules defining the algebra) from data.

Lately, these techniques have been applied to generative models. Indeed, state-of-the-art generative models usually comprise tens of million parameters and are often employed with multidimensional inputs such as color images or multichannel audio signals \cite{Brock2019LargeSG, Patashnik2021StyleCLIP}. The quaternion-valued variational autoencoder and the family of quaternion generative adversarial networks have demonstrated to obtain comparable performance while reducing the storage memory amount due to the parameters reduction \cite{GrassucciICASSP2021, Grassucci2021Entropy, GrassucciQGAN2021, Qgan2021Sfikas}. Encouraged by these results, we propose to expolit novel PHNNs methods to define a more advanced generative model for image-to-image translation.

\section{Quaternion and Parameterized Hypercomplex StarGANv2 Networks}
\label{sec:method}

In this Section, we present the Quaternion StarGANv2 and the Parameterized Hypercomplex StarGANv2 (PHStarGANv2), we describe QNNs and PHNNs involved to build such models and training losses.

\subsection{Models}

We rebuild the real-valued StarGANv2 \cite{Choi2020StarGAN2} as a quaternion model first, and then as a parameterized hypercomplex one to operate in any user-defined hypercomplex domain by easily setting the hyperparameter $n$. Both the quaternion model and the parameterized one are composed of four different networks.

The \textbf{generator network} (G) takes an input image $\x$ and a style code $s$ and translates $\x$ according to $s$ producing a new sample. It is composed of PHC residual blocks (quaternion convolutional blocks for quaternion model), instance normalization and adaptive instance normalization. We carefully redefine the latter in hypercomplex domains, as Section~\ref{sec:qinst} shows more in detail.

\noindent A \textbf{mapping network} (M), instead, takes care of generating a style code $s$ from a sampled latent vector $\mathbf{z}$ and a random domain $y$. It is built by interleaving PHM layers and ReLU activation function. As before, for the quaternion version, we employ quaternion layers instead of PHM ones.

\noindent Third, a \textbf{style encoder network} (S) extracts the style code $s$ from a reference image $\x$. Similar to the generator, the encoder comprises several convolutional residual blocks that are built by PHC layers for PHStarGANv2 models or by Hamilton-based convolutions for quaternion one. However, differently from the generator, the style encoder ends up with a stack of PHM/quaternion layers.

\noindent Finally, the \textbf{discriminator network} (D) is a multi-branch binary classifier that learns to distinguish whether the image $\x$ is a real image of the given domain $y$ or a fake one. This network stacks various residual blocks similar to the already-defined ones with a final fully-connected branch for each domain.

On one hand, the Quaternion StarGANv2 defines these networks in the quaternion domain, involving quaternion operations and layers, thus operating in a pre-defined hypercomplex domain. On the other hand, our PHStarGANv2 is free to run in different domains, thus reproducing the quaternion one setting $n=4$, the complex one with $n=2$ or processing images padding any additional channel by employing $n=3$.

Therefore, we propose two different approaches to perform image-to-image translation in hypercomplex domains. We explore the task in the quaternion domain to leverage the Hamilton product properties and to investigate the model performance with a rigid pre-defined algebra rule as backbone. This method ensures the largest memory saving. We introduce more flexibility thanks to parameterized hypercomplex approaches that guarantee high performance while giving the possibility of choosing the amount of parameters reduction or memory saving and the domain in which the model operates.

\subsection{Losses}

In the following, we expound the losses involved to build the structure of the proposed PHStarGANv2. As the original StarGANv2 \cite{Choi2020StarGAN2}, we employ two approaches for training, both of them based on the same equations. Firstly, we perform a \textit{latent-guided} analysis, thus generating style codes from latent vectors. Secondly, we produce style codes from reference images in the \textit{reference-guided} synthesis. 

The \textbf{adversarial loss} is the known GAN loss \cite{GoodfellowNIPS2014} composed of two binary cross-entropies. Here, it is computed for each branch of the discriminator corresponding to each domain $y$:
\begin{equation}
    \highlight{orange}{$\L_{adv}$} = \Exp_{\x, y}[\log D_y(\x)]+ \Exp_{\x,\tilde{y},\mathbf{z}}[\log(1-D_{\tilde{y}}(G(\x, \tilde{s})))].
\end{equation}
\noindent The \textbf{style reconstruction loss} enforces the generator to employ the style code $\tilde{s}$ when generating images as $G(\x, \tilde{s})$,
\begin{equation}
    \highlight{blue}{$\L_{sty}$} = \Exp_{\x, \tilde{y}, \mathbf{z}}[\left\|\tilde{s}-S_{\tilde{y}}(G(\x, \tilde{s}))\right\|_1].
\end{equation}
\noindent The \textbf{style diversification loss}, instead, forces the generator to explore the image space and to produce diverse and varied images as:
\begin{equation}
    \highlight{green}{$\L_{ds}$} = \Exp_{\x, \tilde{y}, \mathbf{z}_1, \mathbf{z}_2}[\left\|G(\x, \tilde{s}_1)-G(\x, \tilde{s}_2)\right\|_1],
\end{equation}
\noindent whereby $\tilde{s}_1$ and $\tilde{s}_2$, which are the target style codes, are generated by the mapping network from the latent vectors $\mathbf{z}_1$ and $\mathbf{z}_2$. In the reference analysis, these codes are produced from reference images.
The \textbf{preserving source characteristics loss} is a cycle consistency loss aiming at guaranteeing that generated images preserve the domain-invariant features as:
\begin{equation}
    \highlight{purple}{$\L_{cyc}$} = \Exp_{\x, y, \tilde{y}, \mathbf{z}}[\left\|\x - G(G(\x, \tilde{s}),\hat{s})\right\|_1],
\end{equation}
\noindent in which $\hat{s}=S_y(\x)$ is the input $\x$ style code learnt from the style encoder S and $y$ is the true domain of the image. 

\noindent Finally, the \textbf{full loss} is composed by
\begin{equation}
    \underset{G, M, S}{\min}\underset{D}{\max} \; \highlight{orange}{$\L_{adv}$} + \lambda_{sty}\highlight{blue}{$\L_{sty}$} - \lambda_{ds}\highlight{green}{$\L_{ds}$} + \lambda_{cyc}\highlight{purple}{$\L_{cyc}$},
\end{equation}
\noindent with $\lambda_{sty}, \lambda_{ds}, \lambda_{cyc}$ are hyperparameters that balance losses importance.

\section{Hypercomplex Instance Normalizations}
\label{sec:qinst}

Recent models dealing with style losses, such as the original StarGANv2 \cite{Choi2020StarGAN2}, replace batch normalization (BN) with a more suitable normalization technique, namely instance normalization (IN) \cite{8099920}. This method aims at reducing the contrast on the input image in order to make the model focus on the contrast of the reference one only. Therefore, IN normalizes over a single image, differently from BN which normalizes the whole batch. More formally, given a tensor $x \in \mathbb{R}^{N\times C \times H \times W}$ having as element $x_{nchw}$, where $N$ is the batch index, $C$ channels while $H$ and $W$ the spatial dimensions, mean $\mu_{nc}$ and variance $\sigma_{nc}^{2}$ are computed over the latter dimensions and the normalization is applied as
\begin{equation}
\label{eq:in}
y_{nchw}=\gamma \left( \frac{x_{nchw}-\mu_{nc}}{\sqrt{\sigma_{nc}^{2}+\epsilon}} \right) + \beta.
\end{equation}


When dealing with quaternion or hypercomplex inputs, image channels are encapsulated in a single element, thus applying different normalizations per channel, as done in Eq.~\ref{eq:in}, may break relations among components. Therefore, we propose a proper method for computing and employing instance normalization in hypercomplex domains based on the quaternion batch normalization in \cite{VecchiTIT2020}. Suppose to operate in the quaternion domain, the input tensor will be $q \in \mathbb{H}^{N\times C \times H \times W}$ and consequently split in the four quaternion components. Then, the mean $\mu_{nc}$ is still a quaternion and it is computed per component:
\begin{equation}
\begin{split}
\label{eq:mean}
    \mu_{nc}(q) &= \frac{1}{HW} \sum_{h=1}^H \sum_{w=1}^W (q_{0,hw} + q_{1,hw} \ii + q_{2,hw} \ij + q_{3,hw} \ik) \\
    &= \bar{q}_0 + \bar{q}_1 \ii + \bar{q}_2 \ij + \bar{q}_3 \ik.
\end{split}
\end{equation}
The variance is computed in a similar way, however, the per-component values are then averaged and a single variance is then employed for normalization:
\begin{equation}
\begin{split}
\label{eq:var}
    \sigma^{2}_{nc}(q) &= {\frac{1}{HW}}\sum_{h=1}^{H}\sum_{w=1}^{W}(\Delta^{2} q_{0,hw}+\Delta^{2} q_{1,hw} \\
    &+\Delta^{2} q_{2,hw} +\Delta^{2} q_{3,hw}).
\end{split}
\end{equation}
This is an approximation to the optimal variance that is, however, computationally expensive to be calculated due to the particular form of quaternion covariance matrix \cite{GrassucciQGAN2021, MandicSPL2011, TookSIGPRO2011, hoffmann2020algebranets}. Finally, the hypercomplex instance normalization (HIN) can be applied following Eq.~\ref{eq:in} considering that also the shift parameter $\beta = \beta_0 + \beta_1 \ii + \beta_2 \ij + \beta_3 \ik$ is a quaternion.

Following a similar approach, it is possible to redefine also the adaptive instance normalization (AdaIN) \cite{huang2017adain}. As IN, AdaIN was specifically conceived for this kind of applications, and thus it aims at learning shifting and scaling parameters directly from the style. More concretely, AdaIN has no $\gamma$ and $\beta$ parameters since it aligns channel-wise mean and variance of the content image $x$ to the one of the style $y$. For quaternion inputs, we propose to compute normalization statistics of the input $x$ as in Eq.~\ref{eq:mean} and Eq.~\ref{eq:var}. Moreover, we devise to learn style statistics $\mu(y)$ and $\sigma(y)$ through a quaternion (or PHM for hypercomplex inputs) layer to preserve the multidimensional structure of the style input and adaptively learn the parameters. Formally, the hypercomplex AdaIN (HAdaIN) is defined by

\begin{equation}
    \mathrm{HAdaIN}(q,y)=\sigma(y)\left(\frac{q-\mu_{nc}(q)}{\sigma_{nc}(q)}\right)+\mu(y).
\end{equation}

In the experimental Section~\ref{sec:exp}, we empirically test the proposed techniques and show how they increase the generation ability of quaternion and PHStarGANv2.

\section{Weights Initialization for Parameterized Hypercomplex Layers}
\label{sec:init}

\begin{figure}
    \centering
    \includegraphics[width=1\linewidth]{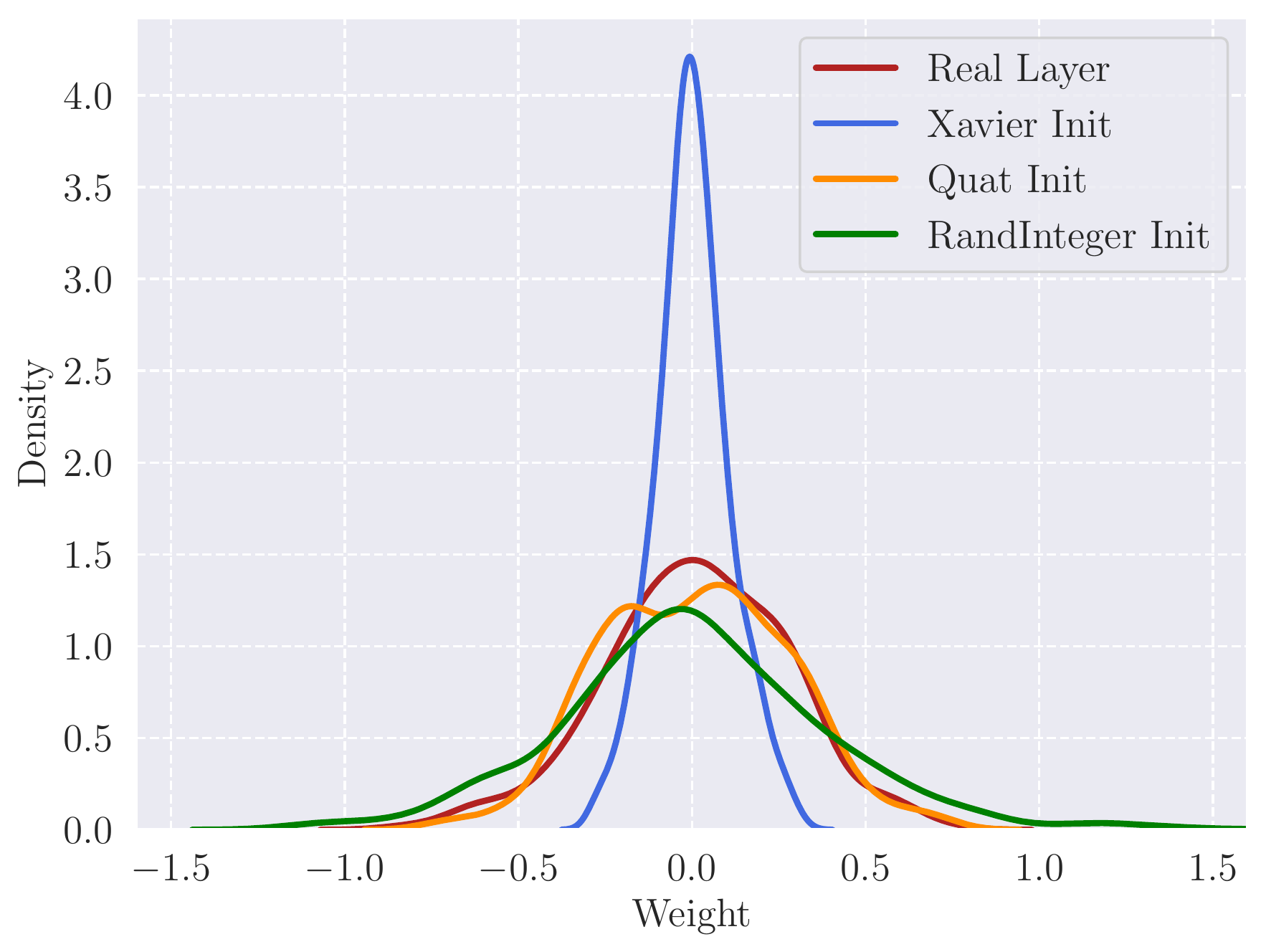}
    \caption{Different initializations for matrices $\mathbf{A}_i$ of a generic PHC layer.}
    \label{fig:init}
\end{figure}

Due to the multidimensional structure of inputs and weights, quaternion and hypercomplex networks initialization has become crucial \cite{RicciardiMLSP2020, ParcolletICLR2019}. While the former has been widely investigated, less attention have been paid to the latest parameterized hypercomplex layers. The original work \cite{grassucci2021lightweight} initializes both $\mathbf{A}_i$ and $\mathbf{F}_i$ with Xavier or Kaiming methods. However, due to Kronecker products among these matrices, these values become smaller and closer to $0$. As a consequence, the resulting weight matrix $\mathbf{H}$ is composed of almost-zero values, as the blue line in Fig.~\ref{fig:init} shows. This may cause training degeneracy \cite{Zhao2021ZerOII, Bachlechner2021ReZeroIA}, undermining the learning procedure. Therefore, since matrices $\mathbf{A}_i$ play the role of \textit{switching on/off} the scalar weights or the filters by defining the algebra rules for multiplication and convolutions, we propose an integer initialization for these PH weights. We select randomly each value in the set $\{-1, 0, 1\}$ meaning that for $0$, the corresponding weight in $\mathbf{F}_i$ is not inserted in the final $\mathbf{H}$, while for $-1$ or $1$, it is considered with minus or plus, respectively. This is inspired to a quaternion-like initialization where the four matrices $\mathbf{A}_i$ are set to reproduce the Hamilton product according to Eq.~\ref{eq:qconv}. This method ensures that the weight matrix $\mathbf{H}$ has a similar distribution to a counterpart real-valued weight matrix, that is the distribution of scalar weights for PHM and filter ones for PHC. We show these comparisons in Fig.~\ref{fig:init}, where the red line is the density of a real-valued convolutional layer initialized with Xavier normal, the yellow one the distribution of a PHC layer initialized rigidly following the Hamilton product in Eq.~\ref{eq:qconv}, thus with $\mathbf{A}_i \in \{-1, 0, 1\}$ while the green line is the $\mathbf{H}$ density of a layer initialized with our RandInteger Init. As it is clear, the final weight matrix is distributed similar to real-valued weights, thus ensuring an analogous behaviour during training, without degeneracy. While the Quat Init can be employed only for $4$D inputs, our method can be generalized to any $n$D inputs and applied with different values of $n$.

\begin{figure*}[t!]
    \centering
    \includegraphics[width=\textwidth]{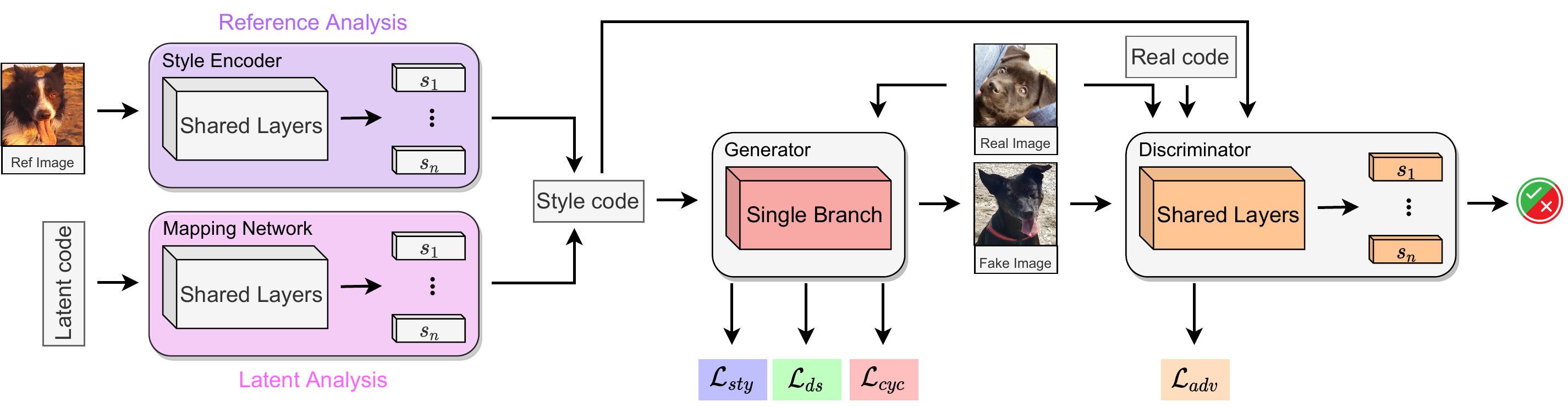}
    \caption{Quaternion StarGANv2 and PHStarGANv2 architecture.}
    \label{fig:arch}
\end{figure*}

\section{Experimental Evaluation}
\label{sec:exp}
In this Section, we conduct a meticulous experimental evaluation of our proposed approaches. We first delineate the experimental setup, including the architectural details and then we report the empirical results.

\subsection{Experimental Setup}

\begin{table*}[t]
\caption{Quantitative comparison on the CelebA-HQ dataset for reference-guided (left) and latent-guided synthesis (right).}
\centering
\label{tab:res_celeba}
\begin{tabular}{l|ccc|cc|cc}
\toprule
 & & & &  \multicolumn{2}{c|}{Reference} & \multicolumn{2}{c}{Latent}  \\
Model & Params & Storage Mem & Savings $\uparrow$ & FID $\downarrow$ & LPIPS $\uparrow$ & FID $\downarrow$ & LPIPS $\uparrow$ \\
\hline
StarGANv2 & 87M & 307MB & 0\% & \textbf{21.24} & 0.24 & 17.16 & 0.25 \\
Quaternion StarGANv2 & 22M & 76MB & \textbf{75\%} & \underline{23.09} & 0.22 & 27.90 & 0.12 \\
PHStarGANv2 $n=3$ & 29M & 137MB & 67\% & 28.11 & \textbf{0.29} & \underline{16.63} & \textbf{0.33} \\
PHStarGANv2 $n=4$ & 22M & 76MB & \textbf{75\%} & 24.33 & \underline{0.27} & \textbf{16.54} & \underline{0.29} \\
\bottomrule
\end{tabular}
\end{table*}

We conduct the experimental evaluation on the CelebA-HQ dataset, which is a high quality version of the CelebA dataset, containing 30000 images at $1024\times1024$ resolution that we rescale at $128\times128$. Moreover, we consider a latent code sampled from a Gaussian distribution with dimension $16$ and a learnt style code of $64$.
In order to have a fair comparison among the networks we consider, we employ the same hyperparameters for real-valued, quaternion and PH models. Except for the batch size, which we set equal to $12$ and not to $8$, the hyperparameters are set as in the original paper \cite{Choi2020StarGAN2}. The losses weights $\lambda_{sty}$ and $\lambda_{cyc}$ are fixed to $1$, while $\lambda_{ds}$ starts from $1$ and decreases after each iteration. Learning is performed via Adam optimizers with $\beta_1=0$ and $\beta_2=0.99$, with a learning rate equal to $10^{-4}$ for all networks except for the mapping one where it is lower $lr=10^{-6}$. We train all networks for 100k iterations.

To delineate models architecture, first we define the blocks we employ in the networks. The first residual block (ResBlock) interleaves convolutional layers, instance normalization and Leaky ReLU activation functions. The second one involves adaptive instance normalization instead of the standard one, therefore we name it AdaResBlock. 

The \textbf{generator network} is built with an initial convolution and with seven ResBlocks, five of them with average pooling for downsampling, seven AdaResBlocks follow, the last five with upsampling. A final refiner convolution is then applied. As the original work \cite{Choi2020StarGAN2}, we involve a pretrained support network which helps in detecting and generating human faces contours and details \cite{wang2020adaptive}.
The \textbf{mapping network} has a simple structure composed by interleaving fully connected (FC) layers and ReLU activations. Four of these layers are shared among the domains, while a four-layer branch is created for each domain.
The \textbf{style encoder} has six shared ResBlocks with a final convolution and a fully connected layer for each domain, whose output is the learnt style code.
Finally, the \textbf{discriminator network} has a similar structure of the style encoder. However, in this case final layers output the probability of an image from the given domain to be real or generated. We test different configurations for the multi-branch last layer and we notice that adding a real-valued FC layer increases the performance, as Table~\ref{tab:improv_FID} shows. Each hyperparameter in these networks are set equal to the original paper, including number of filters, kernel size, and stride, among others \cite{Choi2020StarGAN2}.

The training procedure is depicted in Fig.~\ref{fig:arch}, in which the generator and discriminator networks are employed both for reference (purple) and for latent analysis (pink). The generator gets a real image and a style code and translates the image according to the style code. The discriminator distinguishes between fake and real images given the domain.


To assess the performance of our approaches we compute two objective metrics. First, the Fréchet Inception Distance (FID) measures how much the real and generated distributions are far from each other \cite{NIPS2017_8a1d6947}. Since we ideally want the distributions to be equal, the lower is the FID value, the better is the generation of the model. Second, the Learned Perceptual Image Patch Similarity (LPIPS) \cite{zhang2018unreasonable} measures the diversity of generated images. In this case, the higher is the LPIPS value, the more diverse are the images among each others, meaning a better result.

\subsection{Experimental Results}
To be consistent with the original StarGANv2 evaluation, we replicate both the reference guided analysis and latent guided one. In the former, the style code is extracted by the style encoder from a reference image. Instead, in the latter the style code is learnt by the mapping network starting from a sampled latent code.

\begin{figure}[t]
    \centering
    \includegraphics[width=1\linewidth]{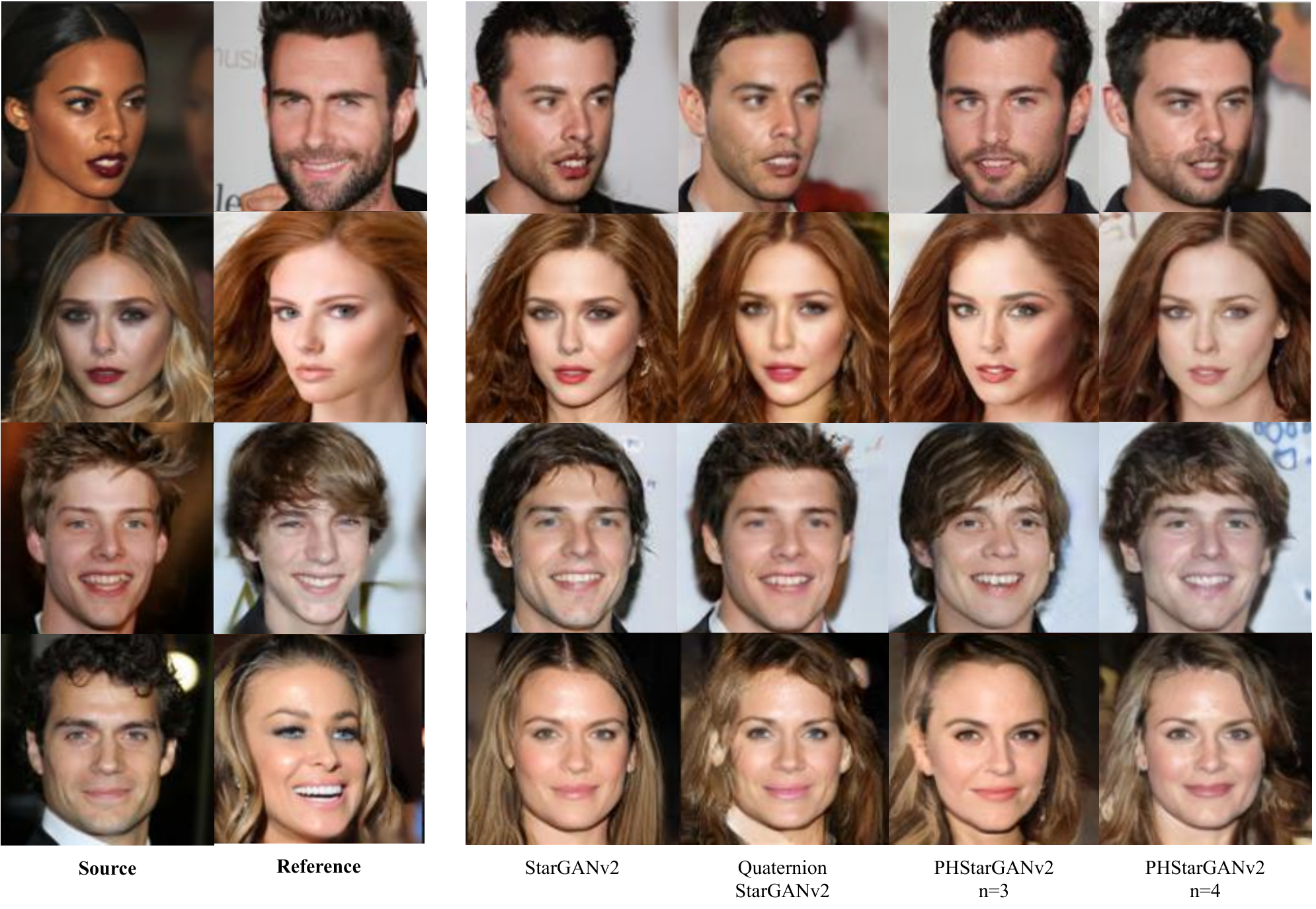}
    \caption{Reference analysis on CelebA-HQ dataset. Source image in the first column, then reference image in the second one. Generated samples from StarGANv2, Quaternion StarGANv2, PHStarGANv2 $n=3$ and PHStarGANv2 $n=4$ follow.}
    \label{fig:reference_celeba}
    \vspace{-0.02cm}
\end{figure}

\begin{figure}[t]
    \centering
    \includegraphics[width=0.85\linewidth]{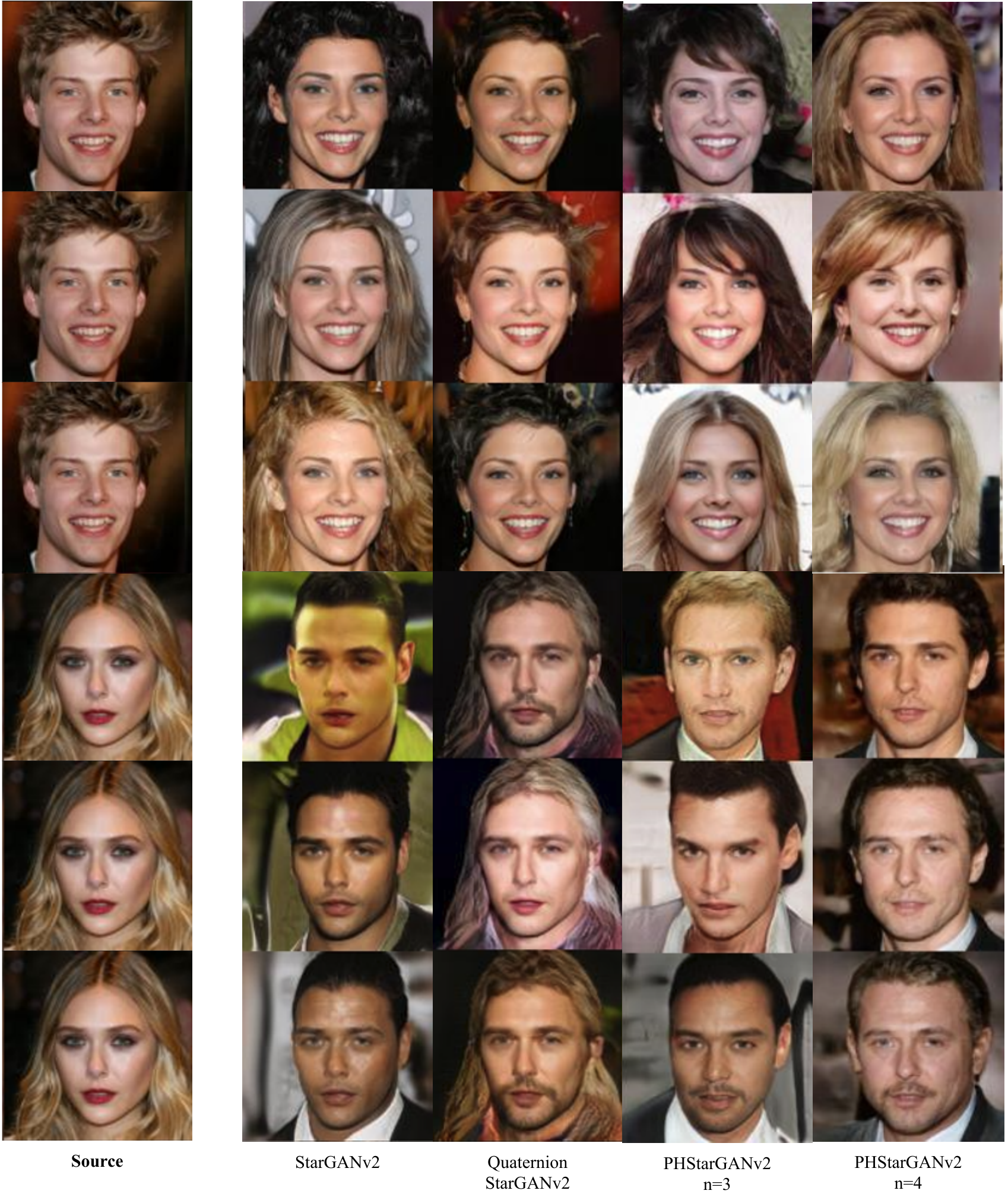}
    \caption{Latent analysis on CelebA-HQ dataset. The source image is in the first column, followed by generated samples from StarGANv2, Quaternion StarGANv2, PHStarGANv2 $n=4$ and PHStarGANv2 $n=3$.}
    \label{fig:latent_celeba}
\end{figure}

Figure~\ref{fig:reference_celeba} displays results for reference analysis, where the style encoder learns to extract the style (hairstyle, color, beard, mouth, eyebrows, among others) from the reference image. The first column of the figure is the source image, while the second one represents the reference. Samples from StarGANv2, Quaternion StarGANv2, PHStarGANv2 $n=3$ and PHStarGANv2 $n=4$ are then showed. While the overall quality of samples is good at a human eye, the PH models better grasp the style from the reference, as it can be seen from the beard of the first row, or from the hairstyle of the third one and from the shape of the face in the last one. The generation quality and samples diversity are measured through objective metrics in Table~\ref{tab:res_celeba} (first column for reference). PHStarGANv2 versions always obtain a better LPIPS scores proving the previous qualitative inspection. Figure~\ref{fig:latent_celeba}, instead, shows samples from the latent analysis. Here, the style is learnt by the mapping network from the sampled latent code. Even in this test, generated samples from PHStarGANv2 models show a better injection of the style code into the source image, thus displaying a more interesting translation while preserving the input pose. Samples from Quaternion StarGANv2 are less diverse and of a lower quality, proving that the high flexibility of PH layers gain advantages with respect to the rigid structure of QNNs. Moreover, PH samples show an improved quality (for example, the last three rows) and human faces are better defined with respect to the real-valued StarGANv2. These results are confirmed by FID and LPIPS results in Table~\ref{tab:res_celeba}, in which the proposed approach outperforms StarGANv2 in each metric considered. More interestingly, these improved results are obtained with a consistent parameters reduction, equal to $-67\%$ employing $n=3$ and to $-75\%$ for the $n=4$ model. This translates in a crucial storage memory saving for checkpoints and inference, as can be seen in Table~\ref{tab:res_celeba}. The choice between different $n$ values can be then left as a user choice or a device-guided one, whether needing an improved diversity generation or a greater storage memory saving.

\subsection{Hypercomplex Instance Normalization Results and Architectural Choices}
In the following, we demonstrate the effectiveness of our hypercomplex instance normalization and we empirically justify some architectural choices. For these experiments we consider as baseline a PHStarGANv2 $n=4$ and real-valued instance normalization, then we compute FID score both for reference and for latent analysis. We then perform the same experiments involving the proposed hypercomplex instance normalization with fixed and with adaptive parameters, noting a great improvement in the FID score (second row of Table~\ref{tab:improv_FID}), proving that our method can help the generation process in hypercomplex domains. Finally, we add a final real-valued fully connected layer to the discriminator network to output the scalar decision value that further slightly increases the FID.
\begin{table}[]
\centering
\caption{FID Improvements with architectural changes (10K iter).}
\label{tab:improv_FID}
\begin{tabular}{l|cc}
\toprule Method & FID Ref $\downarrow$ & FID Lat $\downarrow$ \\ \hline
PHStarGANv2 $n=4$ & 46.36 & 48.73 \\
+ HIN \& HAdaIN & \underline{28.89} & \underline{29.79} \\
+ Last FC layer discriminator & \textbf{27.76} & \textbf{27.77} \\
\bottomrule
\end{tabular}
\end{table}

\subsection{Weights Inizialization Results}

In the following, we report experiments with different initializations for the matrices $\mathbf{A}_i$ in Eq.~\ref{eq:phc} of parameterized hypercomplex layers. We perform three different tests on the CelebA-HQ dataset and we measure performance with FID and LPIPS scores that are reported in Table~\ref{tab:init}. We fix $n=4$ for the PHStarGANv2 and we initialize $\mathbf{A}_i$ with the original method \cite{grassucci2021lightweight}, following the rigid structure of the quaternion product in Eq.~\ref{eq:qconv}, and with our proposed approach with random integers in $\{-1, 0, 1\}$. With the first initialization (Xavier Init) the PHStarGANv2 can not train some weights due to their values close to $0$ and produces almost flat losses. Employing the quaternion initialization (Quat Init), instead, results drastically improve, proving that an integer initialization for matrices $\mathbf{A}_i$ is a good choice. However, produced samples are sometimes blurred and the LPIPS score is very low. Our RanInteger Init, instead, generates high quality images both visually and as measured by FID score, while gaining the best LPIPS values in latent and reference synthesis.
\begin{table}[]
\centering
\caption{Initializations of $\mathbf{A}_i$ in PH layers in PHStarGANv2 $n=4$.}
\label{tab:init}
\resizebox{\linewidth}{!}{
\begin{tabular}{l|cc|cc|c}
\toprule
 & \multicolumn{2}{c|}{Reference} & \multicolumn{2}{c|}{Latent} & \\
 Method & FID $\downarrow$ & LPIPS $\uparrow$ & FID $\downarrow$ & LPIPS $\uparrow$ & Note \\ \hline
Xavier Init \cite{grassucci2021lightweight} & 169.05 & \underline{0.14} & 218.62 & 0.00 & Flat loss \\
Quat Init & \textbf{19.10} & 0.11 & \underline{18.30} & \underline{0.13} & Blurred \\
RandInteger Init & \underline{24.33} & \textbf{0.27} & \textbf{16.54} & \textbf{0.29} & Good \\
\bottomrule
\end{tabular}}
\end{table}

\subsection{Additional I2I Experiments}

\begin{figure}[h!]
    \centering
    \includegraphics[width=1\linewidth]{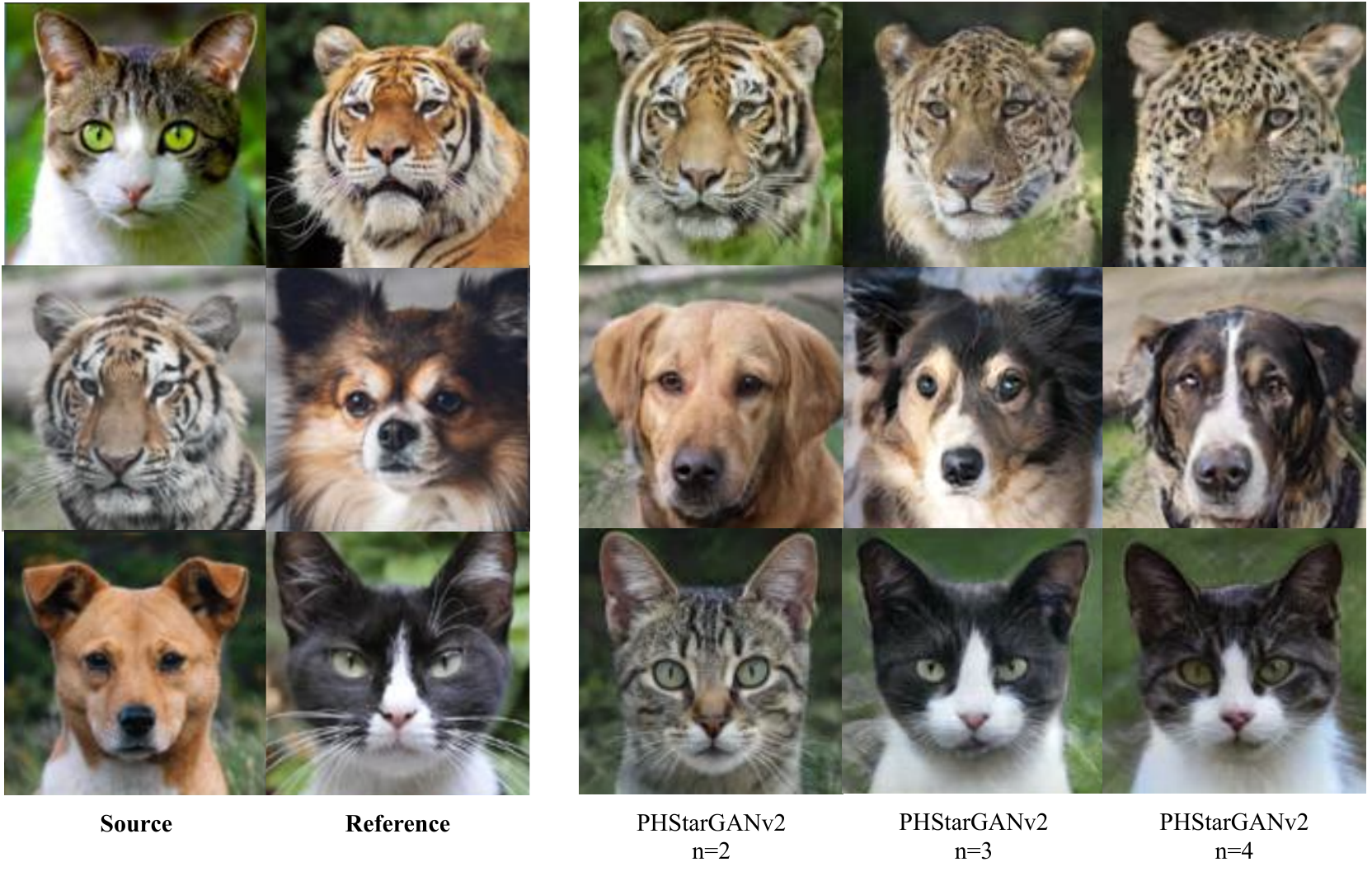}
    \caption{Reference analysis on AFHQ for PHStarGANv2 with $n=2,3,4$.}
    \label{fig:reference_afhq}
\end{figure}

\begin{figure}[h!]
    \centering
    \includegraphics[width=0.83\linewidth]{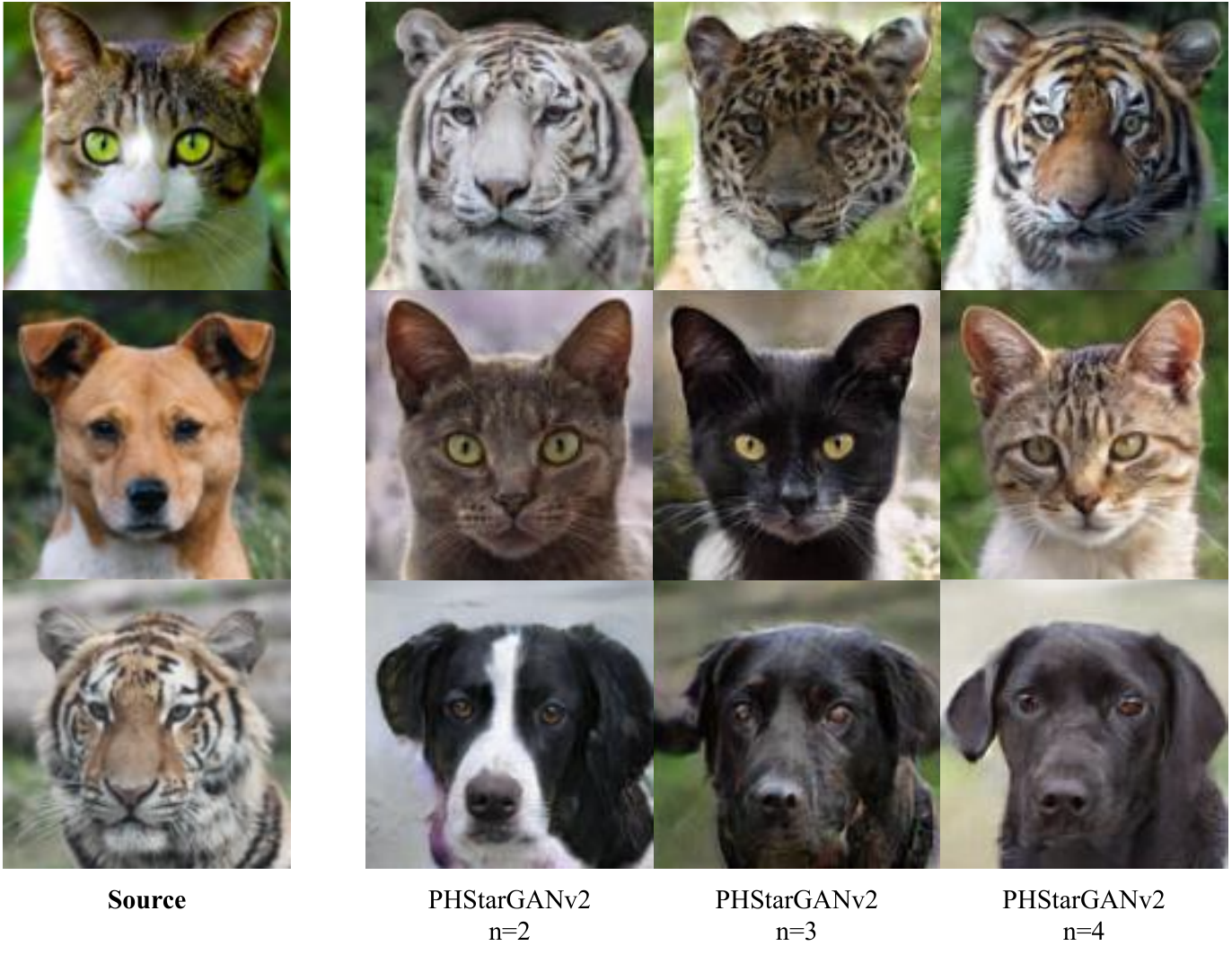}
    \caption{Latent analysis on AFHQ for PHStarGANv2 with $n=2,3,4$.}
    \label{fig:latent_afhq}
\end{figure}

We perform additional experiments to investigate the generation ability of our approach with different values of the hyperparameter $n$ in different benchmarks. For this purpose, we consider the AFHQ dataset containing 15000 samples with aligned animal faces and three different domains \cite{Choi2020StarGAN2}. As for CelebA-HQ, we conduct both reference and latent analysis and keep the same hyperparameters except for $\lambda_{ds}$ equal to $2$ and for $n$ which we test equal to $2,3$ and $4$. Moreover, we discard the support network since it works for human faces only. Figure~\ref{fig:reference_afhq} shows results from the former synthesis, where generated samples preserve source pose while translating attributes from the reference image. As for previous experiments, our PHStarGANv2 is capable of learning the proper style from the reference and then of injecting it in the generator network to perform the domain translation. Indeed, generated samples maintain pose and structure of the input image while modifying animal attributes similarly to the reference (for instance, the ears of the last row). As well, our method demonstrates good I2I translation abilities in latent analysis too, whichever the hypercomplex domain we choose, as Fig.~\ref{fig:latent_afhq} reports. Therefore, we demonstrate how our approach is flexible to operate in different hypercomplex domains and on diverse I2I benchmarks with multiple translation domains.

\section{Conclusion}
\label{sec:conc}

In this paper, we present a quaternion and various hypercomplex approaches for image-to-image translation (I2I) tasks. We propose to exploit quaternion and hypercomplex algebra for lightweight and more reproducible I2I models able to preserve relations among image channels thus grasping additional information from the input. Moreover, we conceive specific normalization modules for hypercomplex domains and a novel initialization for parameterized hypercomplex neural networks. With an empirical evaluation on multiple benchmarks, we demonstrate that our approaches achieve improved results both qualitatively and quantitatively while saving up to $75\%$ parameters and storage memory for inference.

\bibliographystyle{IEEEtran}
\bibliography{HI2IT.bib}

\begin{thebibliography}{10}
\providecommand{\url}[1]{#1}
\csname url@samestyle\endcsname
\providecommand{\newblock}{\relax}
\providecommand{\bibinfo}[2]{#2}
\providecommand{\BIBentrySTDinterwordspacing}{\spaceskip=0pt\relax}
\providecommand{\BIBentryALTinterwordstretchfactor}{4}
\providecommand{\BIBentryALTinterwordspacing}{\spaceskip=\fontdimen2\font plus
\BIBentryALTinterwordstretchfactor\fontdimen3\font minus
  \fontdimen4\font\relax}
\providecommand{\BIBforeignlanguage}[2]{{%
\expandafter\ifx\csname l@#1\endcsname\relax
\typeout{** WARNING: IEEEtran.bst: No hyphenation pattern has been}%
\typeout{** loaded for the language `#1'. Using the pattern for}%
\typeout{** the default language instead.}%
\else
\language=\csname l@#1\endcsname
\fi
#2}}
\providecommand{\BIBdecl}{\relax}
\BIBdecl

\bibitem{Choi2020StarGAN2}
Y.~Choi, Y.~Uh, J.~Yoo, and J.~W. Ha, ``Star{GAN} v2: Diverse image synthesis
  for multiple domains,'' in \emph{{IEEE/CVF} Conf. on Computer Vision and
  Pattern Recognition ({CVPR})}, 2020, pp. 8185--8194.

\bibitem{He2019AttGAN}
Z.~He, W.~Zuo, M.~Kan, S.~Shan, and X.~Chen, ``Att{GAN}: Facial attribute
  editing by only changing what you want,'' \emph{{IEEE} Trans. on Image
  Processing}, no.~11, p. 5464–5478, 2019.

\bibitem{Choi2018StarGAN}
Y.~Choi, M.~Choi, M.~Kim, J.~W. Ha, K.~Sunghun, and C.~Jaegul, ``Star{GAN}:
  Unified generative adversarial networks for multi-domain image-to-image
  translation,'' in \emph{{IEEE/CVF} Conf. on Computer Vision and Pattern
  Recognition ({CVPR})}, 2018.

\bibitem{Lin2021TPAMI}
J.~Lin, Z.~Chen, Y.~Xia, T.~Liu, S.~Qin, and J.~Luo, ``Exploring explicit
  domain supervision for latent space disentanglement in unpaired
  image-to-image translation,'' \emph{{IEEE} Trans. on Pattern Analysis and
  Machine Intelligence}, no.~4, pp. 1254--1266, 2021.

\bibitem{Almahairi2018CycleGAN}
A.~Almahairi, S.~Rajeshwar, A.~Sordoni, P.~Bachman, and A.~Courville,
  ``Augmented {C}ycle{GAN}: Learning many-to-many mappings from unpaired
  data,'' in \emph{Int. Conf. on Machine Learning ({ICML})}, vol.~80, 2018, pp.
  195--204.

\bibitem{Zhang2020CrossDomainCL}
P.~Zhang, B.~Zhang, D.~Chen, L.~Yuan, and F.~Wen, ``Cross-domain correspondence
  learning for exemplar-based image translation,'' \emph{{IEEE/CVF} Conf. on
  Computer Vision and Pattern Recognition ({CVPR})}, pp. 5142--5152, 2020.

\bibitem{Kim2020UGATIT}
J.~Kim, M.~Kim, H.~Kang, and K.~Lee, ``{U-GAT-IT:} unsupervised generative
  attentional networks with adaptive layer-instance normalization for
  image-to-image translation,'' in \emph{Int. Conf. on Learning Representations
  ({ICLR})}, 2020.

\bibitem{Lee2020drit}
H.-Y. Lee, H.-Y. Tseng, Q.~Mao, H.-Y. Huang, Y.-D. Lu, M.~Singh, and M.-H.
  Yang, ``Drit++: Diverse image-to-image translation via disentangled
  representations,'' \emph{Int. Journal of Computer Vision}, pp. 1--16, 2020.

\bibitem{Park2019SPADE}
T.~Park, M.~Y. Liu, T.~C. Wang, and J.~Y. Zhu, ``Semantic image synthesis with
  spatially-adaptive normalization,'' in \emph{{IEEE/CVF} Conf. on Computer
  Vision and Pattern Recognition ({CVPR})}, 2019.

\bibitem{Chen2020Dist}
H.~Chen, Y.~Wang, H.~Shu, C.~Wen, C.~Xu, B.~Shi, C.~Xu, and C.~Xu, ``Distilling
  portable generative adversarial networks for image translation,'' in
  \emph{{AAAI} Conference on Artificial Intelligence}, vol.~34, no.~4, 2020,
  pp. 3585--3592.

\bibitem{Wu2021StyleSpace}
Z.~Wu, D.~Lischinski, and E.~Shechtman, ``Style{S}pace analysis: Disentangled
  controls for {S}tyle{GAN} image generation,'' in \emph{{IEEE/CVF} Conf. on
  Computer Vision and Pattern Recognition ({CVPR})}, 2021, pp.
  12\,863--12\,872.

\bibitem{chong2021gans}
M.~J. Chong and D.~Forsyth, ``{GAN}s {N}' {R}oses: Stable, controllable,
  diverse image to image translation (works for videos too!),'' \emph{ArXiv
  preprint: arXiv:2106.06561}, 2021.

\bibitem{karras2021aliasfree}
T.~Karras, M.~Aittala, S.~Laine, E.~Härkönen, J.~Hellsten, J.~Lehtinen, and
  T.~Aila, ``Alias-free generative adversarial networks,'' in \emph{Advances in
  Neural Information Processing Systems ({NeurIPS})}, 2021.

\bibitem{alaluf2022times}
Y.~Alaluf, O.~Patashnik, Z.~Wu, A.~Zamir, E.~Shechtman, D.~Lischinski, and
  D.~Cohen-Or, ``Third time's the charm? image and video editing with
  {StyleGAN}3,'' \emph{ArXiv preprint: arXiv:2201.13433}, 2022.

\bibitem{GoodfellowNIPS2014}
I.~J. Goodfellow, J.~Pouget-Abadie, M.~Mirza, B.~Xu, D.~Warde-Farley, S.~Ozair,
  A.~Courville, and Y.~Bengio, ``Generative adversarial nets,'' in
  \emph{Advances in Neural Information Process. Systems ({NIPS})},
  vol.~2.\hskip 1em plus 0.5em minus 0.4em\relax Cambridge, MA, USA: MIT Press,
  2014, pp. 2672--2680.

\bibitem{karras2020analyzing}
T.~Karras, S.~Laine, M.~Aittala, J.~Hellsten, J.~Lehtinen, and T.~Aila,
  ``Analyzing and improving the image quality of {S}tyle{GAN},'' \emph{IEEE
  Conf. on Computer Vision and Pattern Recognition ({CVPR})}, 2020.

\bibitem{Brock2019LargeSG}
A.~Brock, J.~Donahue, and K.~Simonyan, ``Large scale {GAN} training for high
  fidelity natural image synthesis,'' in \emph{Int. Conf. on Learning
  Representation ({ICLR})}, 2019.

\bibitem{schonfeld2021unet}
E.~Schönfeld, B.~Schiele, and A.~Khoreva, ``A {U}-{N}et based discriminator
  for generative adversarial networks,'' in \emph{{IEEE/CVF} Conf. on Computer
  Vision and Pattern Recognition ({CVPR})}, 2020, pp. 8207--8216.

\bibitem{ParcolletAIR2019}
T.~Parcollet, M.~Morchid, and G.~Linar\`es, ``A survey of quaternion neural
  networks,'' \emph{Artif. Intell. Rev.}, Aug. 2019.

\bibitem{VALLE2021111}
M.~E. Valle and R.~A. Lobo, ``Hypercomplex-valued recurrent correlation neural
  networks,'' \emph{Neurocomputing}, vol. 432, pp. 111--123, 2021.

\bibitem{ComminielloICASSP2019a}
D.~Comminiello, M.~Lella, S.~Scardapane, and A.~Uncini, ``Quaternion
  convolutional neural networks for detection and localization of 3{D} sound
  events,'' in \emph{{IEEE} Int. Conf. on Acoust., Speech and Signal Process.
  ({ICASSP})}, Brighton, UK, May 2019, pp. 8533--8537.

\bibitem{Brignone2022ISCAS}
C.~Brignone, G.~Mancini, E.~Grassucci, A.~Uncini, and D.~Comminiello,
  ``Efficient sound event localization and detection in the quaternion
  domain,'' in \emph{IEEE Trans. on Circuits and Systems II: Express Briefs},
  vol.~69, no.~5, 2022, pp. 2453--2457.

\bibitem{ParcolletICASSP2019a}
T.~Parcollet, M.~Morchid, and G.~Linar\`es, ``Quaternion convolutional neural
  networks for heterogeneous image processing,'' in \emph{{IEEE} Int. Conf. on
  Acoust., Speech and Signal Process. ({ICASSP})}, Brighton, UK, May 2019, pp.
  8514--8518.

\bibitem{GrassucciQGAN2021}
E.~Grassucci, E.~Cicero, and D.~Comminiello, ``Quaternion generative
  adversarial networks,'' in \emph{Generative Adversarial Learning:
  Architectures and Applications}, R.~Razavi-Far, A.~Ruiz-Garcia, V.~Palade,
  and J.~Schmidhuber, Eds.\hskip 1em plus 0.5em minus 0.4em\relax Cham:
  Springer International Publishing, 2022, pp. 57--86.

\bibitem{GrassucciICASSP2021}
E.~Grassucci, D.~Comminiello, and A.~Uncini, ``A quaternion-valued variational
  autoencoder,'' in \emph{IEEE Int. Conf. on Acoust., Speech and Signal
  Process. ({ICASSP})}, Toronto, Canada, Jun. 2021.

\bibitem{Grassucci2021Entropy}
------, ``An information-theoretic perspective on proper quaternion variational
  autoencoders,'' \emph{Entropy}, vol.~23, no.~7, 2021.

\bibitem{Zhang2021PHM}
A.~Zhang, Y.~Tay, S.~Zhang, A.~Chan, A.~T. Luu, S.~C. Hui, and J.~Fu, ``Beyond
  fully-connected layers with quaternions: Parameterization of hypercomplex
  multiplications with $1/n$ parameters,'' \emph{Int. Conf. on Machine Learning
  ({ICML})}, 2021.

\bibitem{grassucci2021lightweight}
E.~Grassucci, A.~Zhang, and D.~Comminiello, ``{PHNN}s: Lightweight neural
  networks via parameterized hypercomplex convolutions,'' \emph{ArXiv preprint:
  arXiv:2110.04176}, 2021.

\bibitem{mahabadi2021compacter}
R.~K. Mahabadi, J.~Henderson, and S.~Ruder, ``Compacter: Efficient low-rank
  hypercomplex adapter layers,'' \emph{ArXiv preprint: arXiv:2106.04647}, 2021.

\bibitem{le2021parameterized}
T.~Le, M.~Bertolini, F.~Noé, and D.~A. Clevert, ``Parameterized hypercomplex
  graph neural networks for graph classification,'' \emph{ArXiv preprint:
  arXiv:2103.16584}, 2021.

\bibitem{Patashnik2021StyleCLIP}
O.~Patashnik, Z.~Wu, E.~Shechtman, D.~Cohen-Or, and D.~Lischinski,
  ``Style{CLIP}: Text-driven manipulation of {S}tyle{GAN} imagery,'' in
  \emph{{IEEE/CVF} Int. Conf. on Computer Vision ({ICCV})}, 2021, pp.
  2085--2094.

\bibitem{Qgan2021Sfikas}
G.~Sfikas, A.~P. Giotis, G.~Retsinas, and C.~Nikou, ``Quaternion generative
  adversarial networks for inscription detection in {B}yzantine monuments,'' in
  \emph{Pattern Recognition. {ICPR} International Workshops and
  Challenges}.\hskip 1em plus 0.5em minus 0.4em\relax Springer International
  Publishing, 2021, pp. 171--184.

\bibitem{8099920}
D.~Ulyanov, A.~Vedaldi, and V.~Lempitsky, ``Improved texture networks:
  Maximizing quality and diversity in feed-forward stylization and texture
  synthesis,'' in \emph{{IEEE/CVF} Conf. on Computer Vision and Pattern
  Recognition ({CVPR})}, 2017, pp. 4105--4113.

\bibitem{VecchiTIT2020}
R.~Vecchi, S.~Scardapane, D.~Comminiello, and A.~Uncini, ``Compressing
  deep-quaternion neural networks with targeted regularisation,'' \emph{CAAI
  Trans. Intell. Technol.}, vol.~5, no.~3, pp. 172--176, Sep. 2020.

\bibitem{MandicSPL2011}
D.~P. Mandic, C.~Jahanchahi, and C.~Cheong~Took, ``A quaternion gradient
  operator and its applications,'' \emph{{IEEE} Signal Process. Lett.},
  vol.~18, no.~1, pp. 47--50, Jan. 2011.

\bibitem{TookSIGPRO2011}
C.~Cheong~Took and D.~P. Mandic, ``Augmented second-order statistics of
  quaternion random signals,'' \emph{Signal Process.}, vol.~91, no.~2, pp.
  214--224, Feb. 2011.

\bibitem{hoffmann2020algebranets}
J.~Hoffmann, S.~Schmitt, S.~Osindero, K.~Simonyan, and E.~Elsen,
  ``Algebranets,'' \emph{ArXiv preprint: arXiv:2006.07360}, 2020.

\bibitem{huang2017adain}
X.~Huang and S.~Belongie, ``Arbitrary style transfer in real-time with adaptive
  instance normalization,'' in \emph{IEEE Int. Conf. on Computer Vision
  ({ICCV})}, 2017.

\bibitem{RicciardiMLSP2020}
M.~Ricciardi~Celsi, S.~Scardapane, and D.~Comminiello, ``Quaternion neural
  networks for 3{D} sound source localization in reverberant environments,'' in
  \emph{IEEE Int. Workshop on Machine Learning for Signal Process.}, Espoo,
  Finland, Sep. 2020, pp. 1--6.

\bibitem{ParcolletICLR2019}
T.~Parcollet, M.~Ravanelli, M.~Morchid, G.~Linar\`es, C.~Trabelsi, R.~De~Mori,
  and Y.~Bengio, ``Quaternion recurrent neural networks,'' in \emph{Int. Conf.
  on Learning Representations ({ICLR})}, New Orleans, LA, May 2019, pp. 1--19.

\bibitem{Zhao2021ZerOII}
J.~Zhao, F.~Sch{\"a}fer, and A.~Anandkumar, ``Zero initialization: Initializing
  residual networks with only zeros and ones,'' \emph{ArXiv preprint:
  arXiv:2110.12661}, 2021.

\bibitem{Bachlechner2021ReZeroIA}
T.~C. Bachlechner, B.~P. Majumder, H.~H. Mao, G.~Cottrell, and J.~McAuley,
  ``Rezero is all you need: Fast convergence at large depth,'' in \emph{Conf.
  on Uncertainty in Artificial Intelligence ({UAI})}, 2021.

\bibitem{wang2020adaptive}
X.~Wang, L.~Bo, and L.~Fuxin, ``Adaptive wing loss for robust face alignment
  via heatmap regression,'' 2020.

\bibitem{NIPS2017_8a1d6947}
M.~Heusel, H.~Ramsauer, T.~Unterthiner, B.~Nessler, and S.~Hochreiter, ``{GAN}s
  trained by a two time-scale update rule converge to a local nash
  equilibrium,'' in \emph{Advances in Neural Information Processing Systems
  ({NeurIPS})}, vol.~30, 2017.

\bibitem{zhang2018unreasonable}
R.~Zhang, P.~Isola, A.~A. Efros, E.~Shechtman, and O.~Wang, ``The unreasonable
  effectiveness of deep features as a perceptual metric,'' in \emph{{IEEE/CVF}
  Conf. on Computer Vision and Pattern Recognition ({CVPR})}, 2018, pp.
  586--595.

\end{thebibliography}

\end{document}